\documentclass{article} 
\usepackage{iclr2019_conference,times}

\usepackage{amsmath,amsfonts,bm}

\def\eqref#1{equation~\ref{#1}}

\def\1{\bm{1}}

\DeclareMathAlphabet{\mathsfit}{\encodingdefault}{\sfdefault}{m}{sl}
\SetMathAlphabet{\mathsfit}{bold}{\encodingdefault}{\sfdefault}{bx}{n}

\usepackage{hyperref}
\usepackage{url}

\usepackage{float}
\usepackage{caption}
\usepackage{subcaption}
\usepackage{algorithm}
\usepackage[noend]{algpseudocode}
\usepackage[bottom]{footmisc}

\usepackage[utf8]{inputenc} 
\usepackage[T1]{fontenc}    
          
\usepackage{booktabs}       
\usepackage{amsfonts}       
\usepackage{nicefrac}       
\usepackage{microtype}      

\usepackage{amsmath}
\usepackage{xcolor}
\usepackage{amssymb,graphicx}
\usepackage{longtable}

\title{SALSA-TEXT : Self Attentive Latent Space Based Adversarial Text Generation}

\author{Jules Gagnon-Marchand, Hamed Sadeghi, Md. Akmal Haidar \& Mehdi Rezagholizadeh \thanks{} \\
Montreal research Center \\
Huawei Technolgies Canada Co. Ltd. \\
Montreal, Canada \\
\texttt{\{jgagnonmarchand,haamed.sadeghi\}@gmail.com}\\
\texttt{\{md.akmal.haidar, mehdi.rezagholizadeh\}@huawei.com}
}

\iclrfinalcopy 

\begin{document}
\maketitle
\begin{abstract}
Inspired by the success of self attention mechanism and Transformer architecture in sequence transduction and image generation applications, we propose novel self attention-based architectures 
to improve the performance of adversarial latent code-based schemes in text generation.
Adversarial latent code-based text generation has recently gained a lot of attention  due to its promising results. In this paper, we take a step to fortify the architectures used in these setups, specifically AAE and ARAE.
We benchmark two latent code-based methods (AAE and ARAE) designed based on adversarial setups. In our experiments, the Google sentence compression dataset is utilized to compare our method with these methods using various objective and subjective measures. 
The experiments demonstrate the proposed (self) attention-based models outperform the state-of-the-art in adversarial code-based text generation.
\end{abstract}

\section{Introduction}
\noindent 
Text generation is of particular interest in many natural language processing (NLP) applications such as dialogue systems, machine translation, image captioning and text summarization. 
Recent deep learning-based approaches to this problem can be categorized into three classes: auto-regressive or maximum likelihood estimation (MLE)-based, generative adversarial network (GAN)-based and reinforcement learning (RL)-based approaches. 

MLE-based methods (such as \cite{sutskever2014sequence}) model the text (language) as an auto-regressive process, commonly using RNNs. RNNs compactly represent the samples history in the form of recurrent states. In these models, text is generated by predicting next token (character, word, etc) based on the previously generated ones~\citep{graves2013generating}.

One of the main challenges involved with auto-regressive methods is exposure bias~\citep{bengio2015scheduled}. This problem arises due to discrepancy between the training and generation phases. In fact, ground-truth samples from the past are used in training, while past generated ones are used in generation. A number of solutions have been proposed to address this problem by modifying the training procedure including scheduled sampling~\citep{bengio2015scheduled}, Gibbs sampling~\citep{su2018incorporating}, and Professor forcing~\citep{lamb2016professor}.

Over the past few years, researchers have  extensively used GANs~\citep{goodfellow2014generative} as a powerful generative model for text~\citep{yu2017seqgan,che2017maximum}, inspired by the great success in the field of image generation.
GANs are believed to be capable of solving the exposure bias problem in text generation raised from using MLE. The reason is that they solved a similar issue of blurry image generation in MLE-based variational autoencoders (VAEs).
It is belived that the discriminator is able to guide the text generator, through their training exchange, how to generate samples similar to real (training) data. However, there are other challenges involved in GAN-based text generation.



A few of these challenges in text generation are inherent to GANs themselves, such as mode collapse and training instability. The mode collapse problem happens when the adversarially trained generator does not produce diverse texts. These issues can be mitigated by using well-known techniques such as feature matching \citep{zhang2017adversarial}, and entropy regularization~\citep{irl}. Another challenge is due to the discrete nature of text, which causes the generator sampling to be non-differentiable over the categorical distribution of the words. 

In this paper, we take advantage of Transformer self-attention mechanism \citep{transformer} and incorporate it in two state-of-the-art adversarial latent code-based schemes proposed for text generation. More specifically:
\begin{itemize}
\item We incorporate the Transformer structure in the design of encoder and decoder blocks of AAE \citep{aae} and ARAE \citep{arae} setups for text generation.

\item Blocks closely inspired from the Transformer's encoder layers, incorporating self-attention and element-wise fully-connected layers in a residual configuration and with positional encodings, are used along with spectral normalization to propose a novel GAN (both generator and discriminator) structure for AAE and ARAE setups.
\item The performance improvement obtained from the proposed architectures is demonstrated via objective and subjective measures used in extensive experiments.
\end{itemize}

\section{Related Work}

\subsection{Spectral Normalization}
Spectral normalization \citep{spec_norm} is a weight normalization method proposed to stabilize the training of GANs. The authors show that the Lipshitz norm of a neural networks can be bounded by normalizing the spectral norm of layer weight matrices. As opposed to local regularizations used in WGAN-GP, etc., the network-wide spectral regularization stabilizes the GAN training, produces more diverse outputs and results in higher inception scores. We use spectral normalization in our adversarial setups for the same reasons.

\subsection{Attention Models}
In sequence modeling literature, attention was initially proposed by \cite{attention}. It recognizes the fixed-length latent representation of the input sequence as the main performance bottleneck in the seq-to-seq models and proposed using soft-attention in the decoder. Using attention, the decoder can also attend to any desired token in the input sequence besides consuming the compressed representation resulting at the end of encoding operation.

\paragraph{Self-attention} was initially proposed for language inference in \cite{self-attention}. The authors named it as "intra-attention" and showed that their structure can be an effective alternative for LSTMs in the task of natural language inference~\citep{snli}, at the time achieving state of the art performance with much fewer parameters as well as requiring a training time an order of magnitude shorter. Self-attention structures have since been used to set the state of the art in a number of different tasks~\citep{transformer,universal-transformer,qanet,nlutransformerlm, charactertransformerlm}. They drastically reduce the path length between any two sequence inputs, making the learning of long term dependencies much easier~\citep{transformer}. They are considerably easier to parallelize,  reducing the number of operations that are required to be sequential.

Recently, \cite{sagan} applied self attention along with spectral normalization to the task of image generation using GANs. 
It showed by visualization that using attention, the generator can attend to far neighborhoods of any shape rather than close-by fixed-shape ones at each level in a hierarchical generation.
The authors claim that applying spectral normalization to generator as well as discriminator helps training dynamics (stability).
Similarly, we also adopt self attention and spectral normalization in our architecture designs.

Transformer ~\citep{transformer} extended the use of self attention mechanism and was proved to be the state-of-the-art in sequence transduction applications such as machine translation. It dispenses convolutional and recurrent layers and relies entirely on attention-only layers and element-wise feed forward layers.



\subsection{latent space-based text generation}

One of the main challenges of the language generation task originates from the discrete nature of text. Similarly to generating other discrete tokens, the back propagation of error through argmax operator is not well-defined. 
To address this problem, various approaches have been proposed in the literature including continuous approximation of discrete sampling~\citep{gulrajani2017improved,gumbel}, using policy gradient from reinforcement learning~\citep{first-rl,irl}, etc.
One of the most successful solutions is based on autoencoders with continuous latent spaces (i.e. latent code-based methods). Various training setups have been proposed for training these autoencoders including adversarial~\citep{arae} and variational~\citep{vae-controlled} setups.

A recent  paper~\citep{cifka} performs a thorough review of the state-of-the-art latent code-based text generation methods. It studies the performance of a number of code-based text generation schemes and uses a unified rigorous evaluation protocol to evaluate them. We got inspired by their evaluation protocol to demonstrate the strength of our self attention-based approach in the context. They use a broad set of measures to perform a comprehensive study. We adopt forward and reverse perplexity as well as BLEU from their objective measures and fluency from the subjective ones.

\subsection{Adversarial Text Generation}
In this section, we briefly explain two prominent baseline methods using adversarial latent code-based generation techniques and present the technical details in  Section \ref{sec:main}.

\subsubsection{AAE}
Adversarial autoencoder (AAE)~\citep{makhzani2015adversarial} proposes an adversarial setup to train probabilistic autoencoders. It matches the aggregated posterior of the encoder output (latent codes) to an arbitrary distribution that can be easily sampled from. Although authors demonstrate the applications of their setup in semi-supervised learning, style and content disentanglement, etc, AAE decoder can be effectively used as a generative model, converting samples of the arbitrary distribution (noise) to real-like outputs. 
From application perspective, authors only evaluated AAE performance in vision-related applications. In this paper, we tailor AAE for text generation, following guidelines proposed by \cite{cifka} and incorporate self attention and Transformer as novel parts in the model.

\subsubsection{ARAE}
The adversarially regularized autoencoder (ARAE)~\citep{kim2017adversarially} learns an autoencoder with continuous contracted codes that highly correlate with discrete inputs. That is, similar inputs get encoded (mapped) to nearby continuous codes. 
ARAE aims at exploiting GAN’s ability to force the generator to output continuous codes corresponding to the code space obtained from encoding the real text data.
By matching the outputs of generator and encoder, ARAE provides an implicit latent code GAN that serves as a generative model for decoding text.

\section{Self Attentive latent code-based models}
\label{sec:main}

\begin{figure}[htb]
\centering
\includegraphics[scale=0.45]{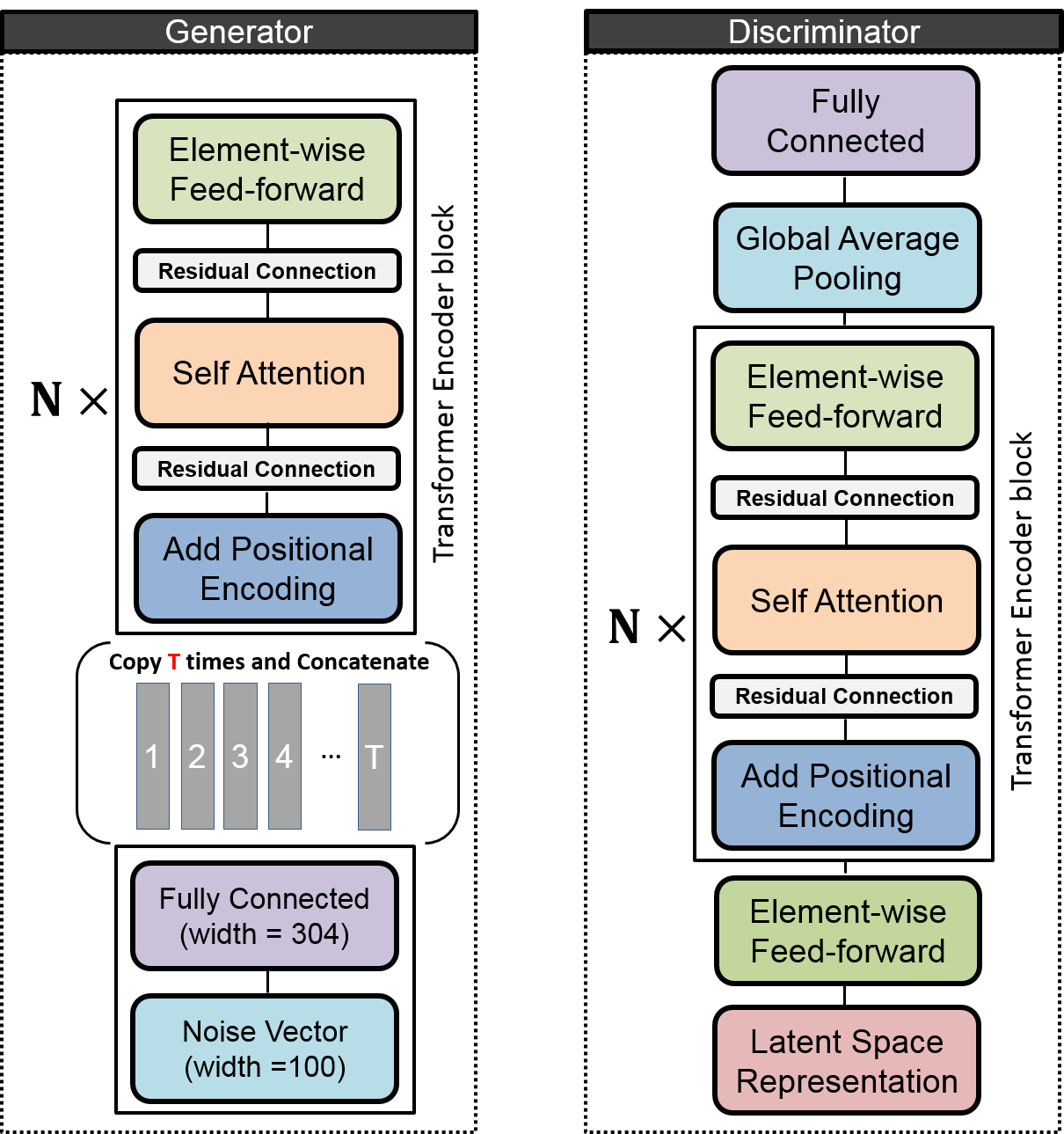}
\caption{SALSA-TEXT generator and discriminator architecture designed using Transformer encoder structure}
\label{fig:gan}
\end{figure}

In this section, we explain the details of our self attention-based models following the ARAE and AAE setups proposed in \cite{cifka}. These setups have shown comparable to the state-of-the-art results in text generation.
We select similar setups to provide fair comparisons and report the best techniques/parameters based on our experiments.

In our architectures, Transformer~\citep{transformer} is used in designing all autoencoders. In both encoder and decoder, we use three blocks of Transformer. `Block' and `layer' names are used, respectively, instead of `layer' and `sub-layer' in the original paper.

Layer normalization\citep{layer-norm} is applied on every layer (multi-head attention, masked multi-head attention and feed forward layers) within each Transformer block. Multi-head attentions have eight heads and embedding layers are of size 304 (a multiple of eight). Similarly to \cite{transformer}, positional encoding is used at the very first layer of the encoder and decoder. The dimensions and encoding place were found empirically for the best objective and subjective performance.

For GAN structures, i.e. the generator and discriminator architectures, we use modified Transformer encoder layers combined with spectral normalization, as depicted in Fig. \ref{fig:gan} ($N = 3$). 
As in the regular transformer blocks, all connections are residual. Inspired by spectral normalization successes in the GAN-based image generation, especially proved in SAGAN ~\citep{sagan}, we apply it to the weights of the discriminator and the generator in our network. We did not find layer normalization (used in original Transformer) to be useful, when applied along with spectral normalization in the generator and discriminator architectures. Hence, only use spectral normalization in our GAN structures.


\subsection{Adversarial Techniques}
We use self attention-based structures in two well-known adversarial setups (\cite{aae} and \cite{arae}).

\paragraph{AAE}

\begin{figure}[!htb]
\centering
\includegraphics[scale=0.6]{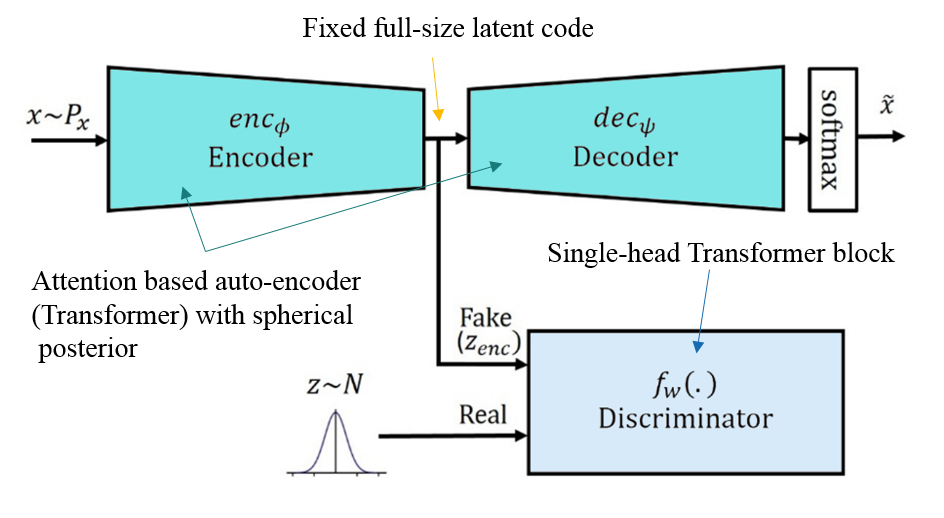}
\caption{SALSA-AAE for text generation. As explained in the figure, Transformer-based encoder, decoder and discriminator (with self attention) architectures are used. Decoder uses fixed-size full codes.}
\label{fig:AAE}
\end{figure}

We use the AAE-SPH setup used in \cite{cifka}. It is based on the original setup proposed in \cite{aae}. The discriminator forces encoder outputs to follow a uniform distribution on the unit sphere. Similarly to \cite{aae}, a two-phase training is used, where there is regular alternation between minimizing reconstruction and adversarial (regularization) costs. The trade-off factor ($\lambda$) between reconstruction and adversarial costs is $20$ (as in \cite{cifka}). All over the encoder, decoder and discriminator, input and attention heads are dropped with a probability of $0.1$.
The general architecture and the proposed (self) attention-based changes are depicted in Fig. \ref{fig:AAE}.

\begin{figure}[!htb]
\centering
\includegraphics[scale=0.6]{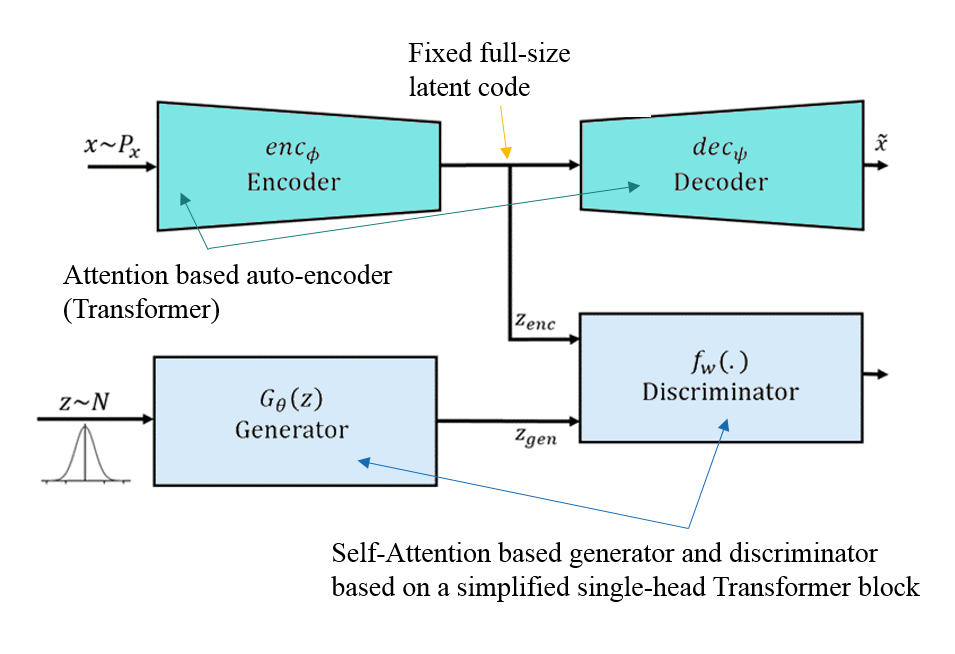}
\caption{SALSA-ARAE for text generation. Similarly to SALSA-AAE, all blocks are Transformer-based and decoder uses fixed-size full codes. Generator and discriminator comprise of self-attention layers.}
\label{fig:ARAE}
\end{figure}

\paragraph{ARAE}
We use the original setup from \cite{arae} with fixed-size full codes as inputs to the decoder. Inside the encoder and decoder, word and attention head dropout is performed with a probability of $0.1$ and a maximum of $3$-word shift is applied to input words. 
The general architecture and the proposed (self) attention-based changes are depicted in Fig. \ref{fig:ARAE}.

\section{Experiments}
We study the performance of our self attentive (SALSA) architectures and compare it with that of the code-based setups studied in \cite{cifka}.

\subsection{Experimental Setup}
The performance of the models is evaluated in sentence generation (sampling), on the Google Sentence Compression (GSC) dataset \footnote{https://github.com/google-research-datasets/sentence-compression} (as in \cite{cifka}). Training on this dataset is very challenging as the sentences are relatively long (average of $24.8$ words) and diverse in terms of content, grammar, etc.
GSC comprises $200,000$ training and $9,995$ test sentences.
For all the trained models, we use Google's SentencePiece \footnote{\url{https://github.com/google/sentencepiece}} tokenizer using byte-pair encoding (BPE~\citep{bpe}) as in \cite{cifka}. 

We filter the dataset to only include sentences with a maximum of $50$ PBE tokens. This only lowers the average number of words per sentence and total number of sentences to $23.1$ and $183739$, respectively in the training set. The test dataset is also reduced to $9254$ lines with an average of $22.7$ words per sentence. Samples of generated sentences from all models is listed in Section \ref{sec:samples}.

The input noise to the generator is of size $100$ (as in \cite{cifka}). We upsample the noise to the embedding size of 304 by using a fully connected layer. The same upsampled noise is copied a number of times equal to the maximum number of steps in the sentence. We use $T = 50$ times in our experiments. The noise is then fed to the generator, where positional encodings are added to each step. The previously mentioned fully connected layer also serves to allow the model to learn to protect the information of the positional encodings from the noise. Positional encodings are also added at the start of each transformer encoder block. As we use fixed size sequences, the attention depth is always fixed ($T$ bpe).
Positional encodings are also added to the input of each transformer encoder block, inside of the generator.

\begin{table}[t]	
  \caption{Samples from models trained on GSC dataset}
  \label{tab:google-samples}
  \small
  \centering
  \begin{tabular}{p{1.5cm}|p{12cm}}
    \toprule
    Model &  Sample sentences \\
    \midrule
    AAE      & $-$ on Thursday for an undisclosed amount of US Senate. \\
    & $-$ .com is launching a new album on ``The Idol's. \\
    & $-$ . \\
    & $-$ Tuesday for a US dollar in the US, according to the US Department. \\
    & $-$ US US markets on Wednesday as US stocks rose to a US dollar, dealers said. \\
    & $-$ The US will open its first-day visit to next year. \\

  \midrule
    SALSA-AAE      & $-$ The world's largest car maker, said it will buy back a new US\$4 million fine for the first time in three years. \\
    & $-$ London, May 6 A 48-year-old man has been charged with raping a woman in the face of livelihood on Tuesday. \\
    & $-$ In a bid to save money, the US government's most expensive land. \\
    & $-$ The Governors of the city of Caterpillar, who was in talks with the world's most expensive officials. \\
    & $-$ Israel has launched a new website that would help the Gaza Strip, and the first such incident in a deadly attack on the West Bank and the United States. \\
    & $-$ Harrington has been found guilty of two counts of driving under the influence of a semi-final at a local court. \\

    \midrule
    
    ARAE      & $-$ A man has been arrested and charged with sexual assault on an alleged assault and assault for allegedly assaulting him to death and a child. \\
    & $-$ A man is accused of stealing a child in her own home case of her husband. \\
    & $-$ A man, who was accused of killing the two-year-old girl in connection with several of them who died from injuries last week. \\
    & $-$ A man is facing charges of sexual assault for allegedly assaulting a woman and her wife. \\
    & $-$ A man accused of killing the man in his home is being sentenced to life and is expected to be on the way. \\
    & $-$ A man has been arrested in connection with her death and then sex with her husband, who was in critical condition and will be released on Monday. \\
    
    \midrule
    
    SALSA-ARAE      & $-$ Former PSV Ethanolve said he has ``two and the title of his contract,'' the first-round pick of the season. \\
    & $-$ The Queen will present a ``Curprone of the Donegal Plan'' in a new biopic of the forthcoming musical, and ``Kalways,'' the school said in a statement. \\
    & $-$ This week between the ``Daily Show'' flights, a year after the airing, a day before the launch of a summer vacation. \\
    & $-$ However, the Myssenon, is planning to open its first wedding, with a move that will include a number of its customers, but they are not planning to sell their services. \\
    & $-$ The Dallas Morning News reported that a Houston man is in ``a new, a state that is leaving the Houston Rockets. \\
    & $-$ The Dallas Morning News Corp. said it is to open a subsidiary of Houston, the largest newspaper and its staff, to be in the city. \\
    
    \bottomrule
  \end{tabular}
\end{table}

\subsection{Evaluation metrics}
\label{sec:metrics}

We use various objective and subjective measures to evaluate the models. As objective measures, we use BLEU~\citep{bleu}, Self-BLEU~\citep{texygen}, forward and reverse perplexity.

\textbf{BLEU}~\citep{bleu} is a widely used metric to compute the similarity of a set of generated sentences with a reference dataset. The results are described in Table \ref{tab:BLEU}.

\textbf{Self BLEU}~\citep{texygen} (Table \ref{tab:SELFBLEU}) is a measure of diversity for generated texts.  In Self-BLEU, for each generated sentence, we compute the BLEU using the sentence as hypothesis and the rest of the generated sentences as the reference. When averaged over all the references, it gives us a measure of how diverse the sentences are. Lower Self-BLEU scores are better, as high BLEU scores indicate great similarity.

In \textbf{Perplexity} evaluation (Table \ref{tab:PPL}), the goal is to measure the individual quality of the sentences generated. We train an LSTM language model on the WMT News $2017$ Dataset \footnote{\url{http://www.statmt.org/wmt17/}} filtered for lines of a maximum of $50$ BPE tokens (a total of $200000$ sentences). The perplexity of the language model is computed over $100000$ generated sentences for each model. 

\textbf{Reverse perplexity} evaluation (Table \ref{tab:PPL}) aims to measure variety of the generated sentences. For each model, we train an LSTM-based language model based on $100000$ generated sentences, and then evaluate the perplexity on the GSC test dataset, filtered to a maximum length of $50$ BPE.
Diverse generated sentences that cover dataset to a good extent would result in better (lower) reverse perplexity measures resulting from the trained LSTM network (language model).

For the \textbf{subjective} evaluation (Table \ref{tab:HE}), we use Amazon mechanical Turk \footnote{https://www.mturk.com/} online platform.
$18$ sentences are sampled from each model, i.e. a total of $162$ sentences.
We assign $81$ randomly selected sentences to $50$ native English speakers (among Mechanical Turk Masters with hit approval ratings greater than $75$\%). The remaining $81$ are assigned to another group of $50$ people with the same qualifications. Each person was asked to evaluate the assigned $81$ sentences in one and a half hours.
In the evaluation, the $5$-point Likart scale is used to measure grammaticality, semantic consistency and overall (Fluency).  The overall reflects both grammar and semantic consistency in addition to other human-specific factors. Hence, it is a good representative of "Fluency" measure used in \cite{cifka}.

\begin{table}[htb]
\centering
{\bf Objective and subjective evaluation of the studied models on the GSC dataset}
\caption {BLEU} \label{tab:BLEU} 
\begin{tabular}{l|cccc|}
\cline{2-5}
&\multicolumn{1}{l|}{AAE} & \multicolumn{1}{l|}{SALSA-AAE} & \multicolumn{1}{l|}{ARAE} & \multicolumn{1}{l|}{SALSA-ARAE} \\ \hline
\multicolumn{1}{|l|}{BLEU-1} & 0.671 & {\bf 0.905} & 0.924 & 0.865 \\ \cline{1-1}
\multicolumn{1}{|l|}{BLEU-2} & 0.500 & {\bf0.638} & 0.698 & 0.585 \\ \cline{1-1}
\multicolumn{1}{|l|}{BLEU-3} & 0.279 & {\bf0.367} & 0.402 & 0.294 \\ \cline{1-1}
\multicolumn{1}{|l|}{BLEU-4} & 0.149 & {\bf0.192} & 0.212 & 0.137 \\ \cline{1-1}
\multicolumn{1}{|l|}{BLEU-5} & 0.086 & {\bf0.101} & 0.116 & 0.071 \\ \hline
\end{tabular}

\bigskip

\caption{Self-BLEU} \label{tab:SELFBLEU} 
\begin{tabular}{l|cccc|}
\cline{2-5}
 & \multicolumn{1}{l|}{AAE} & \multicolumn{1}{l|}{SALSA-AAE} & \multicolumn{1}{l|}{ARAE} & \multicolumn{1}{l|}{SALSA-ARAE} \\ \hline
\multicolumn{1}{|l|}{Self BLEU-1} & 0.950 & {\bf0.897} & 0.973 & {\bf0.896} \\ \cline{1-1}
\multicolumn{1}{|l|}{Self BLEU-2} & 0.759 & {\bf0.738} & 0.921 & {\bf0.731} \\ \cline{1-1}
\multicolumn{1}{|l|}{Self BLEU-3} & 0.604 & {\bf0.573} & 0.843 & {\bf0.549} \\ \cline{1-1}
\multicolumn{1}{|l|}{Self BLEU-4} & 0.412 & {\bf0.410} & 0.751 & {\bf0.367} \\ \cline{1-1}
\multicolumn{1}{|l|}{Self BLEU-5} & 0.270 & {\bf0.275} & 0.653 & {\bf0.226} \\ \hline
\end{tabular}

\bigskip
\caption {Perplexity} \label{tab:PPL} 
\begin{tabular}{c|cccc|}
\cline{2-5}
& \multicolumn{1}{l|}{AAE} & \multicolumn{1}{l|}{SALSA-AAE} & \multicolumn{1}{l|}{ARAE} & \multicolumn{1}{l|}{SALSA-ARAE} \\ \hline
\multicolumn{1}{|c|}{Reverse perplexity} & 10309 &  {\bf822} &  8857 & {\bf1008}\\ \cline{1-1}
\multicolumn{1}{|c|}{Perplexity} & 88 &   {\bf61} &    37 &      106\\ \cline{1-1}
\hline
\end{tabular}
\end{table}

\begin{table}[htb]
\centering
\caption {Human Evaluations}
\label{tab:HE} 
\begin{tabular}{l|cccc|}
\cline{2-5}
\multicolumn{1}{c|}{}  & \multicolumn{1}{c|}{AAE} & \multicolumn{1}{c|}{SALSA-AAE} &
\multicolumn{1}{c|}{ARAE} & \multicolumn{1}{c|}{SALSA-ARAE}\\ \hline
\multicolumn{1}{|l|}{Grammaticality} & 2.756 & {\bf3.09} & 2.980 & 2.898\\ \cline{1-1}
\multicolumn{1}{|l|}{Semantic consistency} & 2.575 & {\bf2.597} & 2.856 & 2.617 \\ \cline{1-1}
\multicolumn{1}{|l|}{Fluency} & 2.604 & {\bf2.700} & 2.851 & 2.652\\ \cline{1-1}
\hline
\end{tabular}
\end{table}
\subsection{Samples of generated sentences}
\label{sec:samples}
In Table \ref{tab:google-samples}, we list six generated sentences for each model.
As seen, AAE generates rather short sentences, while the corresponding SALSA version (SALSA-AAE) has alleviated the issue to a good extent. Finally, ARAE suffers from extreme  mode collapse as opposed to its SALSA counterpart.

\subsection{Results and Discussion}

The results of objective and subjective evaluations are presented in Tables \ref{tab:BLEU} to \ref{tab:HE}. As seen, the proposed self attention-based (SALSA) architectures consistently outperform the non-attention-based benchmarks in terms of diversity (measured by reverse perplexity). Moreover, they often show better performance in terms of output quality (measured by BLEU, self BLEU, preplexity and human evaluations) on the long and complicated sentences of the GSC dataset. 
 
As seen in the generated samples (Table \ref{tab:google-samples}), human evaluation (Table \ref{tab:HE}) and objective metrics (Tables \ref{tab:BLEU} to \ref{tab:PPL}), the original AAE and ARAE setups perform very poorly on GSC with long sentences. With reverse perplexities of over $8000$ and high self-BLEU scores close to $0.9$, they suffer from a high level of mode collapse (repeated sentences). 

Human evaluations do not account for lack of diversity. The reason is humans are presented with a number of shuffled sentences and asked to evaluate them independently (without knowing which sentence coming from which model). Hence, in our experiments for the original AAE and ARAE, a model can generate similar sentences (maybe due to mode collapse) and still receives high subjective scores.

It seems that, in our experiments, the original ARAE model suffers from mode collapse. We can see that it has slightly higher human evaluation scores, but extremely poor diversity metrics, i.e. very high reverse perplexity and self-BLEU scores. It can also be seen in the randomly selected generated sentences (Table \ref{tab:google-samples}), where all the sentences start with "A man" and invariably mention he is being arrested or accused of grievous crimes. This is likely because the sentences in the GSC dataset are long and that their structure is elaborate. SALSA-ARAE on the other hand reliably produces sentences of quality  with great diversity. 

SALSA-AAE has both considerably higher individual quality metrics than the original AAE and much better diversity metrics. It is the strongest pure adversarial text model. As seen in Table \ref{tab:HE}, SALSA-AAE provides the best grammaticality, semantic consistency and Fluency performance.

\section{Conclusion and Future Work}
In this paper, we introduced SALSA-TEXT, a Transformer-based architecture for adversarial code-based text generation. It incorporates self-attention mechanism by utilizing Transformer architecture in autoencoder and GAN setups. Our extensive experiments demonstrate the better performance of our models compared to the state-of-the-art in adversarial code-based text generation (without self-attention). The proposed architectures provide diverse, long and high quality output sentences as confirmed by objective metrics and human evaluations in extensive experiments.

As a future direction, it is beneficial to study the performance of self attention in other text generation methods including variational code-based and reinforcement learning-based approaches.
Another interesting direction is to experiment with deeper Transformer-based autoencoders to better capture the underlying language model and perform unsupervised pre-training isnpired by the success of \cite{charactertransformerlm} and \cite{nlutransformerlm}.

\bibliography{refs}
\bibliographystyle{iclr2019_conference}


\end{document}